# Structuring Bodies of Evidence

**Sandra A. Sandri**
Université Paul Sabatier-IRIT
118, Route de Narbonne
31062 Toulouse France
sandri@irit.fr

## Abstract

In this article we present two ways of structuring bodies of evidence, which allow us to reduce the complexity of the operations usually performed in the framework of evidence theory. The first structure just partitions the focal elements in a body of evidence by their cardinality. With this structure we are able to reduce the complexity on the calculation of the belief functions $Bel$, $Pl$, and $Q$. The other structure proposed here, the Hierarchical Trees, permits us to reduce the complexity of the calculation of $Bel$, $Pl$, and $Q$, as well as of the Dempster's rule of combination in relation to the brute-force algorithm. Both these structures do not require the generation of all the subsets of the reference domain.

## 1 INTRODUCTION

Evidence Theory (Shafer 1976) is a well-known framework for representing uncertainty in Knowledge-based systems. Its use in practical applications is however compromised by the important complexities involved in its manipulation. In particular, the method used to combine the evidence coming from independent sources, known as the Dempster's rule of evidence, may require a complexity of $2^{2n} - 2^{n+1}$ in the worst case, where n stands for the size of the reference domain. The algorithms proposed in the literature to reduce these complexities impose restrictions either on the pieces of evidence themselves, or on the reference domain of the variables modeling them. In the present paper, we are interested to show that some set-theoretical properties underlying bodies of evidence, the set of pieces of evidence, can be used to produce algorithms that are efficient in the situations where the data cannot be restricted.

The text is divided as follows. Section 2 brings some basic notions on Evidence Theory and discusses some of the algorithms implemented within this framework. In Section 3 we present two ways of structuring bodies of evidence : by partitioning the pieces of evidence on the cardinality relation, and by the use of Hierarchical Trees. In Section 4 we propose an algorithm to implement the Dempster's rule using Hierarchical Trees, and in Section 5 we briefly discuss the use of Hierarchical Trees in the Local Propagation of Information. Section 6 brings the conclusion.

## 2 BASIC NOTIONS IN EVIDENCE THEORY

In the framework of Evidence Theory, the information supplied by a source about the actual value of a variable $x$ is encoded in a body of evidence on $\Omega$, where $\Omega$ stands for the set of all the possible values of $x$, called the frame of discernment of $x$. A body of evidence is characterized by a pair $(\mathcal{F}, m)$, where $\mathcal{F}$ is a family of subsets of $\Omega$, ie $\mathcal{F} \subset \mathcal{P}(\Omega)$, and $m$ (called the mass assignement function) is a mapping of $\mathcal{P}(\Omega)$ to the unit interval, such that $m(A) > 0$ iff $A \in \mathcal{F}$, and $\Sigma\{m(A)/A \subset \Omega\} = 1$. Each element $A \in \mathcal{F}$ is called a focal element, and $m(A)$ represents the amount of evidence focused strictly in $A$ itself, and not in any subset of $A$. A body of evidence $(\mathcal{F}, m)$ on $\Omega$ can also be represented by means of any one of the three following set-functions on $\mathcal{P}(\Omega)$ :

$$Bel(A) = \sum\{m(B)/B \subset A, B \neq \emptyset\} \quad (1)$$

$$Pl(A) = \sum\{m(B)/B \cap A \neq \emptyset\} \quad (2)$$

$$Q(A) = \sum\{m(B)/A \subset B\} \quad (3)$$

where $\emptyset$ represents the empty set. The belief function $Bel$ (also called a credibility function) gathers the pieces of information which support $A$. The plausibility function $Pl$ gathers the pieces of information which do not contradict $A$. The commonality function represents to what extent all the elements composing A are plausible (Dubois and Prade 1991). These measures are strongly inter-related ; for instance, the plausibility function $Pl$ can be calculated from the credibility function $Bel$ using

$$Bel(A) = 1 - Pl(\overline{A}) \qquad (4)$$

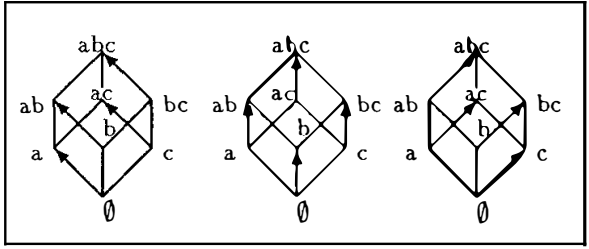

Figure 1: Process of calculation of $Bel$ using Moebius Transforms

Fast algorithms for the calculation of $Bel$, $Pl$, and $Q$ can be found in (Kennes 1990), (Kennes and Smets 1990a), (Kennes and Smets 1990b), and (Thoma 1991). These algorithms are based on Moebius transforms, and are such that each element of $\mathcal{P}(\Omega)$ can be seen as an object that receives and sends information in the form of accumulated evidence. In the calculation of any of the belief functions, each element propagates its accumulated evidence only once. Fig. 1 illustrates the calculation of $Bel$ for all the focal elements in $(\mathcal{F}, m)$, on $\Omega = \{a, b, c\}$.

Dempster's rule of combination is the method used for pooling evidence in the framework of Evidence Theory (Shafer 1976). The combination of two given bodies of evidence $(\mathcal{F}_1, m_1)$ and $(\mathcal{F}_2, m_2)$ on $\Omega$, yields a third body $(\mathcal{F}_3, m_3)$ on $\Omega$, where

$$m_3(A) = \sum \{m_1(B) * m_2(C) / B \cap C = A\} \qquad (5)$$

This result may be unnormalized, i.e. we might have $m_3(\emptyset) > 0$. Originally, the application of Dempster's rule involves a normalization step, in which the mass assigned to each focal element of $(\mathcal{F}_3, m_3)$ is divided by a constant $K = 1 - m_3(\emptyset)$, representing the amount of conflict between the sources. The effective use of this normalization step is however quite controversial (see (Dubois and Prade 1988), (Smets 1988) for discussions over this point).

The use of (5) for the calculation of the Dempster's rule, here called the brute-force strategy, has a complexity of $|\mathcal{F}_1| * |\mathcal{F}_2|$ in terms of focal elements visited, which gets to $2^{2n} - 2^{n+1}$ in the worst case. Drastic reductions on this complexity are however achieved in some restricted situations. Barnett (1981) treated the case in which the evidence is presented in the form of bodies of evidence having two focal elements : a singleton $f \in \Omega$, and its complement in $\Omega$. Gordon and Shortliffe (1985), and Shafer and Logan (1987) treated the case in which all the possible evidence is hierarchical, in such a way as that all the possible focal elements can be arranged as nodes in a tree, where each father-node represents the union of its son-nodes, and all the nodes having a common father are disjoint. Moreover, in this model, all the bodies of evidence consist of only two focal elements : one representing the reference domain $\Omega$, and another that is either a node of the tree, or the complement of one of its nodes. The continuous case has been treated by Strat (1984), with the restriction that all the focal elements should be closed intervals. A general discussion on the complexity of Dempster's rule can be found in (Orponen 1990).

Another way of implementing Dempster rule, here called the $Q$-strategy, consists in calculating the commonality functions $Q_1$ and $Q_2$ for every set $A \in \Omega$, and then using an important property of the commonality function $Q$, namely :

$$Q_3(A) = Q_1(A) * Q_2(A) \qquad (6)$$

The mass assigment function $m_3$ can then be recovered by using (Smets 1988) :

$$m(A) = \sum \{(-1)^{|B|} Q(A \cup B), B \subset \overline{A}\} \qquad (7)$$

The exact number of operations performed on the $Q$-strategy with the direct utilization of formulae (6) and (7) is $3^n + 2^{2n+1} - 2^{n+1}$. However, this value comes down to $(n+1)2^n + n2^{n-1} - n$, when Moebius transforms are used (Kennes and Smets 1990a), and (Kennes 1990). The single inconvenience with this strategy is that it requires that he whole set $\mathcal{P}(\Omega)$ be generated, being thus unapplicable when $\mathcal{P}(\Omega)$ is not enumerable.

## 3 PROPOSAL OF STRUCTURES

We see that efficient algorithms for the calculation of both belief measures and Dempster rule are achieved with restrictions on either the evidence, or on the frame of discernment. Set-theoretical properties underlying the bodies of evidence can however be used to produce efficient algorithms without loss of expressivity.

One of the simplest of such properties is that the focal elements contained in $f \in \mathcal{F}$ are either $f$ itself, or elements having cardinality smaller than $f$. Based on this property, we propose to structure a body of evidence with $\mathcal{V}(\mathcal{F})$, the partition induced on $\mathcal{F}$, when the focal elements are classified by their cardinality. With this structure we are able to reduce the complexity of the calculation of the belief measures $Bel$, $Pl$, and $Q$.

An important characteristic of $\mathcal{P}(\Omega)$ is that it forms a lattice with the relation $\subset$ (in particular, it is this property that underlies the use of Moebius Transforms in Evidence Theory). Since $\mathcal{F} \subset \mathcal{P}(\Omega)$, $\mathcal{F}$ forms an incomplete lattice with $\subset$. We present here a structure, here, called Hierarchical Trees, that uses this property in order to reduce the complexity of the implementation of Dempster's rule, as well as of $Bel$, $Pl$, and $Q$. This structure establishes an hierarchy in a given body of evidence ; each node $f$ in the tree (representing a focal element $A$) is connected to a father-node and to a set of sons-nodes, which are respectively greater or smaller than $f$ in the sense of inclusion. Algorithms based on both of the structures proposed here do not require that the whole set $\mathcal{P}(\Omega)$ be generated, and can thus be employed when $\mathcal{P}(\Omega)$ is not enumerable.

### 3.1 PARTITION $\mathcal{V}(\mathcal{F})$

Let $(\mathcal{F}, m)$ be a body of evidence on $\Omega$, and $A$ and $B$ be two of its focal elements contained in $\Omega$. We can structure $\mathcal{F}$ with the partition $\mathcal{V}(\mathcal{F}) = \{c_i, 1 \leq i \leq |\Omega|\}$, where $A$ $c_i$ $B$, iff $|A| = |B| = i$, ie $A$ and $B$ will belong to the same class if they have the same cardinality. The construction of $\mathcal{V}(\mathcal{F})$ is obviously linear with $|\mathcal{F}|$, and can be done as the input is read.

The credibility function $Bel$ on a set $A \in \mathcal{P}(\Omega), |A| = j$, in a given body of evidence $(\mathcal{F}, m)$, can be calculated from the partition $\mathcal{V}$ corresponding to $\mathcal{F}$ using :

$$Bel(A) = m(A) + \sum_{i=1}^{j-1} \{m(B), B \in c_i, A \supset B\} \quad (8)$$

$Bel(A)$ is calculated by adding its own mass $m(A)$ to the masses of its subsets that can be found lying in the classes below it in $\mathcal{V}$. Each focal element thus visits itself plus the elements with lower cardinality than itself in $\mathcal{V}$. Thus if $|c_i|$ represents the number of elements of cardinality $i$ present in partition $\mathcal{V}$, the total cost of the algorithm is $|\mathcal{F}| + \sum_{i=1}^{n} \sum_{j=1}^{i-1} |c_i||c_j|$. The calculation of the plausibility function $Pl(A)$ is similarly done by using formula (4), with the application of (11) in the calculation of $Bel(\overline{A})$. The total cost of calculating $Pl$ for all the elements of $\mathcal{F}$ is $|\mathcal{F}| + \sum_{i=1}^{n} \sum_{j=1}^{n-i-1} |c_i||c_j|$. The commonality function $Q$ can be calculated using the formula :

$$Q(A) = m(A) + \sum_{i=j+1}^{n} \{m(B), B \in c_i, A \subset B\} \quad (9)$$

leading to a cost of $|\mathcal{F}| + \sum_{i=1}^{n} \sum_{j=i+1}^{n} |c_i||c_j|$, when applied to all the elements in $\mathcal{F}$.

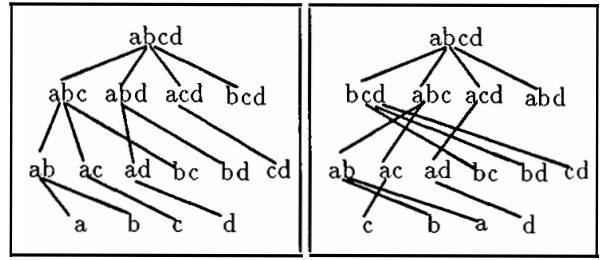

Figure 2: Hierarchical trees on $\mathcal{P}(\Omega)$, $\Omega = \{a, b, c, d\}$

### 3.2 HIERARCHICAL TREES

Let the set $\mathcal{A}(A) = \{B \mid B \in \mathcal{F}, B \supset A, B \neq A\}$, be the set of ancestors of $A$ in $\mathcal{F}$, ie the set of proper supersets of $A$ in $\mathcal{F}$. Similarly, let $\mathcal{D}(A) = \{B \mid B \in \mathcal{F}, A \supset B, B \neq A, B \neq \emptyset\}$, be the set of descendants of $A$ in $\mathcal{F}$, ie the set of all proper subsets of $A$ (except the empty set) in $\mathcal{F}$. A Hierarchical Tree is a structure that relates the focal elements in $\mathcal{F}$ with their respective ancestors and descendants sets. Let $f_i$ be a node representing a set $A_i \in \mathcal{P}(\Omega)$. A Hierarchical Tree $T = (N, E)$ on the body of evidence $(\mathcal{F}, m)$ consists of a set of distinct nodes $N = \{f_i \mid A_i \in \mathcal{F}\}$ and a set of edges $E = \{(f_i, f_j) \mid f_i, f_j \in N, f_i \neq f_j, A_i \supset A_j, \nexists A_i \supset A_k \supset A_j\}$. Set $R = \{f_i \mid \nexists, (f_k, f_i) \in E\}$ denotes the roots of $T$, and $Sons(f) = \{f_i \mid (f, f_i) \in E\}$ denotes the set of sons of a node $f$ in the Hierarchical Tree $T$. A Hierarchical Tree may in fact have several roots, thus constituting a forest. Note however that any forest can be transformed into a tree with the addition of a dummy root node $u$, with $m(u) = 0$, where $u$ represents the union of the focal elements in $\mathcal{F}$. Throughout this paper we suppose that $R$ has a single element. To simplify the notation we make $N = \mathcal{F}$, and $f_i = A_i$.

Hierarchical Trees on $(\mathcal{F}, m)$ can be seen as the structures that can be derived from the incomplete lattice induced on $\mathcal{F}$ with the relation $\subset$ when we extract edges in the lattice, in such a way that each node will have at most one father. Thus, from the same set of focal elements $\mathcal{F}$ several Hierarchical trees $T_k = (\mathcal{F}_k, m_k)$ may be derived. They are equivalent for our purposes, in the sense that if a focal element $A$ is a node $f$ in $T_k$, then all the nodes in the path linking $f$ to a root node belong to $\mathcal{A}(A)$, and all the nodes accessible from $f$ belong to $\mathcal{D}(A)$. Moreover, the father of $f$ in $T_k$ refers to a focal element with the smallest cardinality among the nodes in $\mathcal{A}(A)$. Fig. 2 shows some Hierarchical trees derivable from $\mathcal{F} = \mathcal{P}(\Omega), \Omega = \{a, b, c, d\}$. Note that the trees described in both (Gordon and Shortliffe 1985), and (Shafer and Logan 1987) are particular cases of Hierarchical Trees, in which all the nodes in a given level are disjoint focal elements.

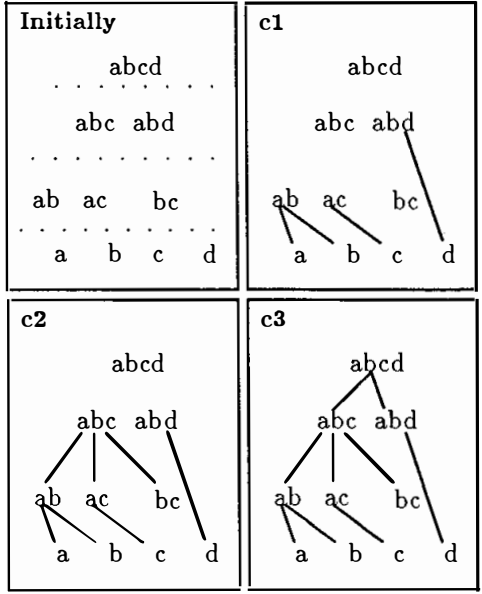

Figure 3: Process of creation of a Hierarchical Tree for $\mathcal{F} = \{abcd, abc, abd, ab, ac, bc, a, b, c, d\}$. Each box refers to the identification of fathers for the nodes in class $c_i$.

### 3.2.1 Hierarchical Trees Construction

In order to construct a Hierarchical Tree for a body of evidence $(\mathcal{F}, m)$, we take partition $\mathcal{V}(\mathcal{F})$ and create to each focal element $A$ with cardinality $i$, a node $f_{ij}$ in $T$. We start the process of identification of the father of $f_{ij}$ by examining the nodes on class $c_{i+1}$. If there is no possible father on class $c_{i+1}$, then $f_{ij}$ is compared with the focal elements on class $c_{i+2}$, and so successively. If no father can be found, then $f_{ij}$ is set as a root node. Fig. 3 illustrates the application of this algorithm for $\mathcal{F} = \{abcd, abc, abd, ab, ac, bc, a, b, c, d\}$). (Fig 3-a shows partition $\mathcal{V}(\mathcal{F})$).

The tree construction in Fig. 3 requires the examination of 13 nodes, and the complete tree on $\mathcal{P}(\Omega)$, with $\Omega = \{a, b, c, d\}$, would require examining 22 nodes. The exact number of the nodes visited in order to construct the worst possible tree with $|\Omega|$ elements is (Sandri and Dugat 1991) : $2^n + 2^{n-3} - 2 + \sum_{i=3}^{n-1} \binom{2i}{i-2} 2^{n-i-1}$.

### 3.2.2 Calculation of Belief Measures

An important characteristic of Hierarchical Trees is that a son-node always "inherits" the ancestors of its father-node. This property can be advantegeously used in the calculation of the $Q$ function, by comparing a Hierarchical Tree on $\mathcal{F}$ with the partition $\mathcal{V}(\mathcal{F})$, as it is seen in the algorithm presented below.

Let $f$ be a node in the tree $T$, $Q(f)$ be the commonality function of $f$, and $Sons(f)$ and $Father(f)$ respectively be the immediate sons and the father of $f$ in $T$. We compute $Q$ recursively for all the nodes in tree $T$ with the application of the following algorithm (initially $f = r$, and $Q(Father(r)) = 0$:

*If node $f$ (of cardinality $i$) has at least one son, it is compared to each element $f_{ij}$ of its own class in $\mathcal{V}$. Otherwise, it is only compared to itself. If $f \cap f_{ij} \neq \emptyset$, we update the masses in $\mathcal{V}$ by transfering the mass on $f_{ij}$ to $f \cap f_{ij}$, by making $m(f \cap f_{ij}) \leftarrow m(f \cap f_{ij}) + m(f_{ij})$, and $m(f_{ij}) \leftarrow 0$, if $f \cap f_{ij} \neq f_{ij}$. $Q(f)$ is computed as $Q(Father(f)) + m(f)$, and the algorithm is successively repeated for all the nodes in $Sons(f)$.*

The comparisons that $f$ effectuates in its own level have the sole objective of transfering the masses of elements that might be ancestors of its sons, to subsets of $f$ (note that if $g \in Sons(f)$, and $l \supset g, then(f \cap l) \supset g)$. Each node visits at least itself, and when it has sons, it visits also all of its class neighbours. The maximal number of nodes visited by this algorithm is $2^{n-1} - n + \frac{1}{2}\binom{2n}{n}$, that is closely bounded by $2^{n-1} - n + \frac{2^{2n-1}}{\sqrt{n\pi}}$ (Sandri and Dugat 1991). For instance, the cost of calculating $Q$ for all $\mathcal{F} = \mathcal{P}(\Omega)$ with $|\Omega| = 5$ is 961 with the usual algorithm, 386 using partition $\mathcal{V}$, and 211 using a Hierarchical Tree : 74 for the tree construction and 137 for the $Q$ algorithm itself (the approximation gives 141 instead of 137 for the $Q$ algorithm).

$Bel$ and $Pl$ can be calculated from the commonality function associated with the complement of the body of evidence, as exposed in (Dubois and Prade 1986). The complement of a body of evidence $(\mathcal{F}, m)$ is $(\neg \mathcal{F}, \overline{m})$, defined as $\forall A \subset \Omega, \overline{m}(A) = m(\overline{A})$, so that $\neg \mathcal{F} = \{\overline{A}/A \in \mathcal{F}\}$. Function $Q$ is defined as :

$$\overline{Q}(\overline{A}) = \sum \{\overline{m}(\overline{B}) / \overline{A} \subset \overline{B}\} \qquad (10)$$

The $Bel$ function is then calculated using

$$Bel(A) = \overline{Q}(\overline{A}) - m(\emptyset) \qquad (11)$$

Finally, function $Pl$ is calculated using (10), (11), and (4). In the worst case, the calculation of $Bel$ and $Pl$ using Hierarchical Trees requires, besides the nodes visited in the calculation of Q, the visit of additional $2^n - 1$ nodes, $|\Omega| = n$, due to the derivation of $\neg \mathcal{F}$.

## 4 DEMPSTER RULE OF COMBINATION

Shafer (1987) comments that an exponential complexity seems to be intrinsic to Dempster's rule. Indeed, the worst case complexity has to be calculated on $|\mathcal{P}(\Omega)|$, since it represents the largest value that $|\mathcal{F}|$ may take. However, using the Q-strategy, the complexity does not decrease when $|\mathcal{F}| << |\mathcal{P}(\Omega)|$, i.e. we are always in a position of the worst case analysis no matter how the evidence is presented. This situation occurs because this strategy requires the generation of the whole set $\mathcal{P}(\Omega)$, and thus deals with more than just the sets $\mathcal{F}_1$, $\mathcal{F}_2$ and $\mathcal{F}_3$ involved in the process. On the other hand, efficient algorithms for the implementation of the brute-force strategy impose restrictions on the evidence. When we recall that bodies of evidence benefit from set-theory properties, it seems natural that there should exist ways of calculating Dempster's rule whose complexity in the mean case depends exclusively on $|\mathcal{F}_1|$ and $|\mathcal{F}_2|$, and that restricts neither the evidence, nor the frame of discernment.

We propose here an algorithm for calculating Dempster's rule using the brute-force strategy that takes advantage of the set-theoretical properties underlying two given bodies of evidence. We divide the process in two phases, the pre-processing phase, and the combination phase in itself.

**Pre-Processing Phase**

Let $u_1$ and $u_2$ respectively be the union of the focal elements of $\mathcal{F}_1$ and $\mathcal{F}_2$. It is obvious that the highest possible focal element of the resulting body of evidence ($\mathcal{F}_3$, $m_3$) is $u_3 = u_1 \cap u_2$. We can thus reduce any of the bodies $\mathcal{F}_k$ by comparing $u_3$ to each of its focal elements $f_{ij}$, and transporting the mass on $f_{ij}$ to $f_{ij} \cap u_3$, ie making $m(f_{ij} \cap u_3) \leftarrow m(f_{ij} \cap u_3) + m(f_{ij})$ and $m(f_{ij}) \leftarrow 0$, if $f_{ij} \cap u_3 \neq f_{ij}$. Note that $\mathcal{F}_k$ remains of the same size in two cases: a) $f_{ij} \cap u_3$ is not an element of $\mathcal{F}_k$, and then $f_{ij} \cap u_3$ will be created as $f_{ij}$ will be eliminated (thus modifying $\mathcal{F}_k$ but not its size), or b) $f_{ij} = f_{ij} \cap u_3$, and then $\mathcal{F}_k$ is not modified. On the other hand, when $f_{ij} \neq f_{ij} \cap u_3$, and $f_{ij} \cap u_3$ belongs already to $\mathcal{F}_k$, the element $f_{ij}$ is simply eliminated, thus reducing the size of $\mathcal{F}_k$.

In the pre-processing phase we compare $u_3$ only to the smallest body of evidence between $\mathcal{F}1$ and $\mathcal{F}2$ (the other body of evidence will be implicitly compared to $u_3$ in the next phase). Thus, if $\mathcal{F}_1$ is the smallest body of evidence, the cost of the pre-processing phase is 1 if $u_1 = u_3$, and $|\mathcal{F}_1|$ otherwise.

**Combination Phase**

Let us suppose that $\mathcal{F}_1$ is the smallest body of evidence, and $T_1$ the Hierarchical Tree for $\mathcal{F}_1$ resulting

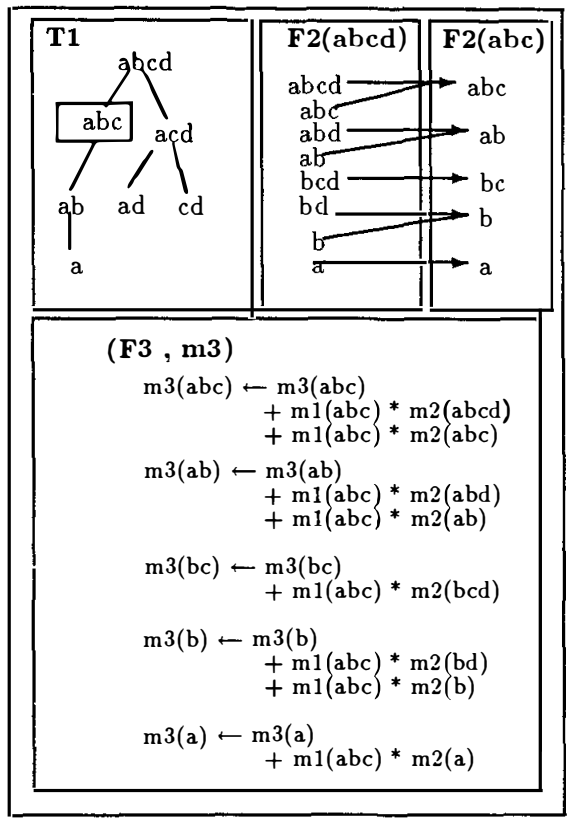

Figure 4: Combination of $abc$ and $\mathcal{F}_2(Father(abc))$

from the construction algorithm as seen in Section 2. Let $f$ be a node in $T_1$, $Father(f)$ the father-node of $f$ in $T_1$, and $\mathcal{F}_2(Father(f))$ be the result of the intersection of $Father(f)$ with all the elements of $\mathcal{F}_2$. We will combine each node in $T_1$ to the focal elements of $\mathcal{F}_2$ by applying the following algorithm, taking the root node as the initial value of $f$ and $\mathcal{F}_2(Father(r))$ as the set $\mathcal{F}_2$ itself:

*Initially $\mathcal{F}_2(f) = \emptyset$. We compare $f$ to each focal element $g$ in $\mathcal{F}_2(Father(f))$. If $f \cap g \neq \emptyset$, we update $m_2(f \cap g)$ in $\mathcal{F}_2(f)$ with $m_2(g)$, and $m_3(f \cap g)$ with $m_1(f) * m_2(g)$. Then we successively apply the algorithm for the son-nodes of $f$ in $T_1$ with $\mathcal{F}_2(f)$ as input.*

Fig. 3 illustrates the combination of node $f = abc$ with $\mathcal{F}_2(Father(abc)) = \{abcd, abc, abd, bcd, ab, bd, a, b\}$.

Note that the set of elements $\mathcal{F}_2(Father(f))$ examined by any node $f$ in $T_1$ is at most the power set $\mathcal{P}(Father(f))$ (the root node examines at most $\mathcal{P}(\Omega)$). The maximal cost of the combination phase is $2^m + 2 * 3^n - 3 * 2^n - 1$, with $n = |u_1|$ and $m = |u_2|$ (Sandri and Dugat 1991). This algorithm turns out to be more efficient the largest is $|\mathcal{F}_1| \times |\mathcal{F}_2|$. For instance, two complete bodies of evidence on $\Omega$, with $|\Omega|=5$, will require visiting 496 nodes (74 for the tree construction, 1 for the pre-processing phase, and 421 for the combination phase), instead of 961 of the brute-force algorithm. This strategy is specially recommended if, after the pre-processing phase, the maximal cost of constructing a Hierarchical Tree and then of combining it, is found to be greater than $|\mathcal{F}_1| \times |\mathcal{F}_2|$. Nevertheless, experimental results show that this strategy is worse than the brute-force algorithm only when $T_1$ is composed exclusively of a root node $r$ and its immediate sons $Sons(r)$. In this case, all the nodes in $T_1$ visit all the focal elements in $\mathcal{F}_2$, and thus the additional cost of the tree construction is not justified.

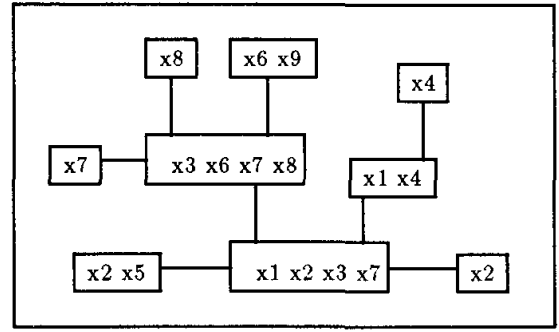

Figure 5: Markov Tree on $\mathcal{X} = \{x_1, x_2, ..., x_9\}$

## 5 LOCAL PROPAGATION OF INFORMATION

Knowledge Bases are usually composed of individualized pieces of data, each of which regarding a cluster of variables, that together model the system knowledge about the world. When uncertainty is modeled within the framework of Evidence Theory, these pieces of data can be characterized by bodies of evidence on $\Omega_{\mathcal{X}}$, where $\mathcal{X} = \{x_1, x_2, ..., x_n\}$ denotes the set of all the variables in the Knowledge Base. Let $G \subset \mathcal{X}$ be a set of variables in $\mathcal{X}$, $\Omega_G = \Omega_{x_{G^1} x_{G^2} \ldots x_{G^g}} = \Omega_{x_{G^1}} \times \Omega_{x_{G^2}} \times \ldots \times \Omega_{x_{G^g}}$, $x_{G^i} \in \mathcal{X}$. Each focal element of a body of evidence $(\mathcal{F}_G, m_G)$ on $\Omega_G$, is then a set of g-uples on $\Omega_{x_{G^1}} \times \Omega_{x_{G^2}} \times \ldots \times \Omega_{x_{G^g}}$. The extension and projection of a body of evidence $(\mathcal{F}, m)$ are done with the application of the usual set-theoretical extension and projection operations on the set $\mathcal{F}$, and on the additive function $m$. $(\mathcal{F}_G{}^{\uparrow J}, m_G{}^{\uparrow J})$ and $(\mathcal{F}_G{}^{\downarrow H}, m_G{}^{\downarrow H})$ respectively denote the extension and projection of $(\mathcal{F}_G, m_G)$ to a higher and lower dimensional space, $H \subset G \subset J \subset \mathcal{X}$.

The belief function for a multi-dimensional variable $S \subset \mathcal{X}$, taking into account all the evidence in the Knowledge Base, can be obtained by first calculating the overall belief function on $\mathcal{X}$, and then marginalizing this overall belief function to $S$. The overall belief function can be obtained by taking all the clusters of variables $G \subset \mathcal{X}$ on $\Omega_G$ present in the Knowledge Base, extending each of them to the highest possible frame of discernment $\Omega_{\mathcal{X}} = \Omega_{x1} \times \Omega_{x2} \times \ldots \times \Omega_{xn}$, and then applying Dempster rule on the resulting set of extended clusters. The cost of the computation of such a procedure is however prohibitive. An alternative is to use the Local Propagation of Information (Shafer and Shenoy 1986), (Shafer and Shenoy 1988a) and (Shafer and Shenoy, 1988b). This strategy requires that the Knowledge Base be partitioned in groups of pieces of knowledge, in such a way as that the variables involved in each group can be arranged as nodes in a Markov Tree (also called a Join Tree in recent literature). In this tree, the nodes are clusters of variables, and the edges are such that, if a variable is contained in two nodes, then it is contained in all the nodes along the path between these two nodes (see Fig. 5).

The process of Local Propagation of Information on a Markov Tree consists on propagating information from the leaf-nodes until the root node, by means of projection/ extension/ combination operations. To obtain the marginal on $S \subset \mathcal{X}$ from the overall belief function, it suffices to propagate information locally on a Markov Tree, setting as root one of the nodes containing $S$.

When a body of evidence $(\mathcal{F}_G, m_G)$ on $\Omega_G$ is extended to a frame of discernment of higher dimension $\Omega_J$, $J \subset \mathcal{X}$, $G \subset J$, it undergoes only a minor changement ; the focal elements change (a focal element $f$ becomes $f \times \Omega_{J-G}$), but the inter-relationships between the focal elements remain the same. Thus there exists a Hierarchical Tree derivable from $(\mathcal{F}_G{}^{\uparrow J}, m_G{}^{\uparrow J})$ whose nodes and edges have a one-to-one correspondance with those in the Hierarchical Tree $T$ constructed for $(\mathcal{F}_G, m_G)$. This tree is in fact the tree that we obtain by changing the label on each node in $T$ with its extension on $\Omega_J$. However, in case of projection of a body of evidence $(\mathcal{F}_G, m_G)$ on $\Omega_G$ to a frame of discernment of lower dimension $\Omega_H$, not only the focal elements will change but also the structure may change to a simpler one, since many focal elements on $\Omega_G$ may be projected on the same focal element on $\Omega_H$. As a consequence, only changing the labels of the focal elements of $T$ does not suffice to generate a Hierarchical Tree on $(\mathcal{F}_G{}^{\downarrow H}, m_G{}^{\downarrow H})$. However, we obtain a Hierarchical Tree $T_H$ on $(\mathcal{F}_G{}^{\downarrow H}, m_G{}^{\downarrow H})$, directly from $T$ if the projection operation on $(\mathcal{F}_G, m_G)$ is performed in the following manner. Let $f$ be a node in $\mathcal{F}_G$, $Father(f)$ the father-node of $f$, and $f^{\downarrow H}$ the projection of $f$ on $\Omega_H$. We start the projection process on the root node, towards the leaf-nodes. Everytime a node $f^{\downarrow H}$ is created in $T_H$, we take as its father the node $Father(f)^{\downarrow H}$ that represents the projection of the father of $f$ on $\Omega_H$, i.e. we create an edge $(Father(f)^{\downarrow H}, f^{\downarrow H})$ in $T_H$. At the end of the process we obtain a Hierarchical Tree $T_H$, that represents the projection of $T$ on $\Omega_H$. More details on the manipulation of Hierarchical Trees in the Local Propagation of Information can be found in (Sandri Dugat 1991).

Thoma (1991) exposes the way Moebius Transforms can be efficiently employed in the application of the $Q$-strategy in the Local Propagation of Information. As in the case of a single variable, the choice between the brute-force or the $Q$-strategy for the combination of the information on a node $G$, depends on whether $\mathcal{P}(\Omega_G)$ is enumerable or not. If there exists a node $G$ in the Markov Tree, such that $\mathcal{P}(\Omega_G)$ is not enumerable then Q-strategy can be used until the process reaches $G$ ; afterwards only the brute-force strategy can be used in the remainder of the propagation process.

## 6  CONCLUSION

The choice between the brute-force and the $Q$-strategy should be guided by the relation between $\mid \mathcal{P}(\Omega) \mid$ and $\mid \mathcal{F}_1 \mid * \mid \mathcal{F}_2 \mid$, when $\mathcal{P}(\Omega)$ can be enumerated. In the case where the use of the brute-force strategy is obligatory, and $\mid \mathcal{F}_1 \mid * \mid \mathcal{F}_2 \mid$ is rather large, structuring the bodies of evidence is the only possible way of reducing the computational load, without the imposition of restrictions on the data.

In this article we presented two ways of structuring a body of evidence ; by partitioning its focal elements by their cardinality, and by constructing Hierarchical Trees, that allows us to take into account the set inter-relationships existing among the focal elements in a body of evidence. We also presented algorithms for calculating belief measures and the Dempster's rule of combination, that do not require the creation of the whole set of possible focal elements $\mathcal{P}(\Omega)$. These structures do not impose any restrictions on the data, and can be easily manipulated in the Local Propagation of Information.

### Acknowledgements

The author is mostly indebted to Vincent Dugat for the help on the calculation of the complexities involved in this work.